\documentclass{article}

\usepackage{microtype}
\usepackage{graphicx}
\usepackage{subfigure}
\usepackage{xspace}
\usepackage{booktabs} 

\usepackage{hyperref}
\usepackage{amsmath,multicol}
\usepackage{subcaption}
\usepackage{graphicx}
\usepackage{caption}
\usepackage{xcolor}
\usepackage{listings}
\usepackage{multirow}
\usepackage{booktabs}
\usepackage{verbatim}
\usepackage{mdframed}

\usepackage{stfloats}

\definecolor{delim}{RGB}{20,105,176}
\definecolor{numb}{RGB}{106, 109, 32}
\definecolor{string}{rgb}{0.64,0.08,0.08}

\colorlet{punct}{red!60!black}
\definecolor{background}{HTML}{EEEEEE}
\definecolor{delim}{RGB}{20,105,176}
\colorlet{numb}{magenta!60!black}
\usepackage{multirow}
\usepackage{colortbl}
\usepackage[T1]{fontenc}
\lstdefinelanguage{json}{
    basicstyle=\scriptsize\ttfamily,
    numbers=none,
    numberstyle=\scriptsize,
    stepnumber=1,
    numbersep=8pt,
    showstringspaces=false,
    breaklines=true,
    frame=lines,
    backgroundcolor=\color{background},
    literate=
     *{0}{{{\color{numb}0}}}{1}
      {1}{{{\color{numb}1}}}{1}
      {2}{{{\color{numb}2}}}{1}
      {3}{{{\color{numb}3}}}{1}
      {4}{{{\color{numb}4}}}{1}
      {5}{{{\color{numb}5}}}{1}
      {6}{{{\color{numb}6}}}{1}
      {7}{{{\color{numb}7}}}{1}
      {8}{{{\color{numb}8}}}{1}
      {9}{{{\color{numb}9}}}{1}
      {:}{{{\color{punct}{:}}}}{1}
      {,}{{{\color{punct}{,}}}}{1}
      {\{}{{{\color{delim}{\{}}}}{1}
      {\}}{{{\color{delim}{\}}}}}{1}
      {[}{{{\color{delim}{[}}}}{1}
      {]}{{{\color{delim}{]}}}}{1},
}

\newcommand{\sys}{\texttt{DroidCall}\xspace}

\usepackage[accepted]{icml2021}

\begin{document}

\twocolumn[
\icmltitle{\sys: A Dataset for LLM-powered Android Intent Invocation}




\begin{icmlauthorlist}
\icmlauthor{Weikai Xie}{to}
\icmlauthor{Li Zhang}{to}
\icmlauthor{Shihe Wang}{to}
\icmlauthor{Rongjie Yi}{to}
\icmlauthor{Mengwei Xu}{to}
\end{icmlauthorlist}

\icmlaffiliation{to}{Beijing University of Posts and Telecommunications (BUPT), China}

\icmlcorrespondingauthor{Mengwei Xu}{mwx@bupt.edu.cn}


\vskip 0.3in
]



\printAffiliations{}

\begin{abstract}

The growing capabilities of large language models in natural language understanding significantly strengthen existing agentic systems.
To power performant on-device mobile agents for better data privacy, we introduce \sys, the first training and testing dataset for accurate Android intent invocation. With a highly flexible and reusable data generation pipeline, we constructed 10k samples in \sys.
Given a task instruction in natural language, small language models such as Qwen2.5-3B and Gemma2-2B fine-tuned with \sys can approach or even surpass the capabilities of GPT-4o for accurate Android intent invocation.
We also provide an end-to-end Android app equipped with these fine-tuned models to demonstrate the Android intent invocation process.
The code and dataset are available at \url{https://github.com/UbiquitousLearning/DroidCall}.

\end{abstract}
\section{Introduction}

The advent of large language models (LLMs) revolutionizes natural language processing, enabling machines to understand and generate human-like language with unprecedented accuracy. In the realm of mobile computing, this advancement presents a significant opportunity for developing intelligent mobile agents~\cite{li2024personal,zhang2024llamatouch,wen2024autodroid,wang2023enabling}.
Specifically, these agents can leverage the rich ecosystem of built-in \texttt{intents} \cite{intent} provided by both the operating system and third-party applications on Android devices.
These intents serve as a fundamental mechanism for inter-app communication and function invocation, such as sending messages, making phone calls, or triggering specific app features.
By harnessing LLMs, mobile agents can interpret diverse and complex user instructions, seamlessly mapping them to the appropriate intents, and therefore automating user interaction with mobile devices.

\begin{figure}[t]
    \centering
    \includegraphics[width=0.93\linewidth]{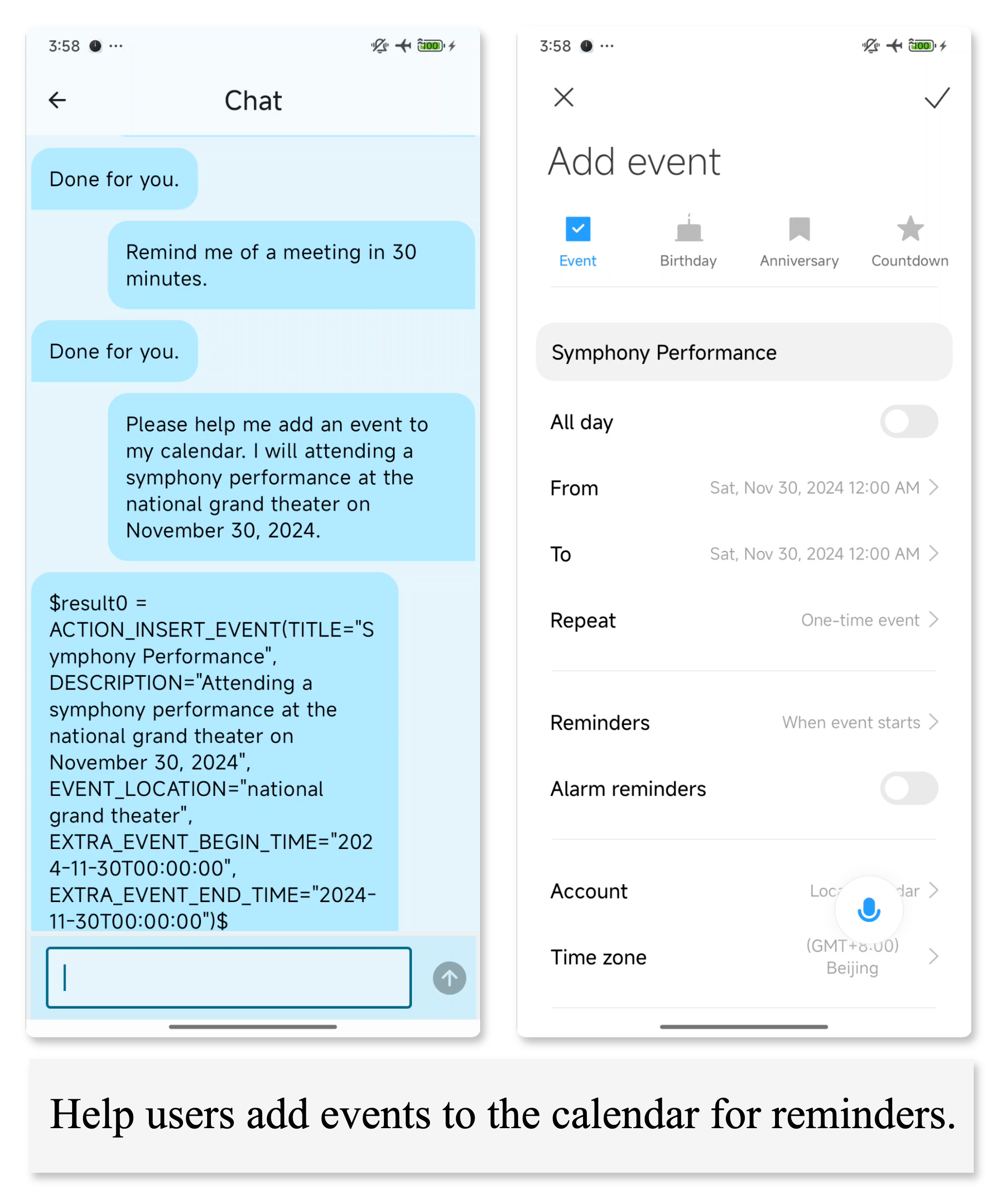}
    \caption{Small language models fine-tuned with \sys have the capability to assist users in completing common tasks such as adding events to the calendar.}
    \label{fig:demo}
\end{figure}

On-device LLMs are necessary for building mobile agents due to privacy and latency constraints~\cite{google-aicore,lu2024small,yin2024llm,xu2023llmcad,yuan2024mobile}.
Since user data are processed locally, sensitive information remains on devices, thereby mitigating risks associated with data transmission over networks.
Moreover, on-device inference eliminates the need for constant internet connectivity.
Various on-device LLM inference optimizations significantly reduce response time~\cite{xu2024survey,yi2023edgemoe,xu2024empowering}, leading to a more responsive and fluid user experience.

However, our investigations reveal a critical challenge: Existing device-affordable LLMs lack the capability of accurate intent invocation.
For example, Llama3.2-1B~\cite{dubey2024llama} only succeeds in 31.5\% and 60.5\% of the tasks in zero-shot and few-shot scenarios, respectively.
This limitation is not due to inherent deficiencies in the models themselves but stems from the absence of specialized datasets tailored for this purpose. Existing LLMs are typically trained on broad datasets that do not encompass the specific language patterns and contextual nuances required for accurate intent invocation. 

To address this gap, we introduce \sys, the first open-sourced, high-quality dataset designed for fine-tuning LLMs for accurate intent invocation on Android devices.
\sys comprises an extensive collection of user instructions paired with their corresponding intents, covering a wide array of functionalities across the system and third-party apps.
We predefined functions encapsulating the process of intent invocation, aiming for the model to learn their invocation for acquiring intent invocation capabilities.
With predefined functions, this pipeline automatically generates data followed by format validation and deduplication to ensure accuracy, relevance, and diversity.
Unlike other similar methods \cite{selfinstruct, alpaca, qin2023toolllm}, our pipeline does not require manual writing of seed data, thus saving a significant amount of labor.

\textbf{Evaluation.} Based on \sys, we fine-tuned a few small language models (SLMs), including PhoneLM-1.5B \cite{yi2024phonelmanefficientcapablesmall}, Qwen2.5-1.5B-Instruct, Qwen2.5-3B-Instruct, Qwen2.5-Coder-1.5B \cite{qwen2.5}, Gemma2-2B-it \cite{gemmateam2024gemma2improvingopen}, Phi-3.5-mini-instruct \cite{abdin2024phi3technicalreporthighly}, MiniCPM3-4B \cite{hu2024minicpmunveilingpotentialsmall}, Llama3.2-1B-Instruct, Llama3.2-3B-Instruct \cite{dubey2024llama}. We demonstrate that by fine-tuning models on \sys, the Android Intent invocation capabilities of these SLMs can be effectively unleashed. Some models can even achieve higher accuracy than GPT-4o using simpler prompts. While prompts for GPT-4o contain an average of 1,367 tokens,  models after fine-tuning, achieve this with an average of just 645 tokens.
The accuracy of using Gemma2-2B improves from 59\% to 85\% after fine-tuned on \sys, while GPT-4o only achieves an accuracy of 77\%.

\textbf{End-to-end demo and open-source.}
We also provide an end-to-end Android demonstration with the fine-tuned models based on mllm~\cite{yi2023mllm}, a fast and lightweight multimodal LLM inference engine, which demonstrates the feasibility of our work.
The demo is illustrated in Figure~\ref{fig:demo}, which can assist users in completing common operations such as composing emails, setting alarms, making phone calls, and so on. \sys is available at \url{https://github.com/UbiquitousLearning/DroidCall}

\section{Related Work}

\subsection{LLM-based Agents}

LLMs have emerged as a significant advancement in the field of artificial intelligence, marking a new era in natural language processing and understanding. Among them, OpenAI's GPT series \cite{achiam2023gpt} has ushered AI into the era LLMs, which have begun to enter the public eye and develop rapidly. Subsequently, numerous open-source LLMs \cite{qwen2, qwen2.5, bai2023qwen, dubey2024llama, liu2024deepseek, zhu2024deepseek, glm2024chatglm} have emerged, gradually approaching and even rivaling the capabilities of GPT-4, which has empowered developers and researchers alike to harness the power of these advanced models to implement a variety of applications. Furthermore, models such as GPT-4V have endowed LLMs with visual capabilities \cite{yang2023gpt4v, lu2024deepseekvl, wang2024qwen2vl, liu2024llava}, enabling them to undertake a broader and more complex array of tasks.

By employing some prompting techniques, such as React \cite{yao2022react}, Plan and Solve \cite{wang2023plan}, ReWOO \cite{xu2023rewoo}, it is possible to guide LLMs in planning for specific tasks. These approaches enable the models to use tools and interact with the external environment, thus enhancing their capabilities to perform more intricate tasks. Based on LLMs and innovative prompting methods, a variety of agents such as AutoGPT \cite{yang2023auto}, MetaGPT \cite{hong2023metagpt} and HuggingGPT \cite{shen2024hugginggpt} which can serve as assistants to humans have emerged. 

\subsection{Mobile Device Control Agents}

Significant efforts have been made in controlling mobile devices using agents.
Early work \cite{venkatesh2022ugif, wang2023enabling, wen2024autodroid} design UI representations to bridge the gap between GUIs and natural language, enabling models to understand mobile screens.
Later, with the advent of multimodal LLMs, agents become capable not only of processing textual inputs but also of receiving images, audio, or video as inputs. This enhancement allows them to perceive the external environment more effectively and accomplish more complex tasks.
Work such as AppAgent \cite{yang2023appagent} and Mobile Agent \cite{wang2024mobile, wang2024mobile2} integrate visual capabilities to implement agents on mobile devices. 

However, most existing agents have certain limitations.
(1) Most of them utilize cloud-side LLMs such as GPT-4. Applications on edge devices prioritize user privacy protection, and implementing edge agents through invoking cloud-side LLMs cannot effectively safeguard user privacy. Additionally, agents cannot be used under poor network conditions. Our work addresses these issues by deploying SLMs on edge devices to control Android devices, which effectively avoids the aforementioned problems. 
(2) Existing agents heavily rely on simulating human actions to operate mobile devices, such as through tap and swipe gestures. In this study, we envision agents directly interfacing with mobile devices through intent invocation as a more efficient and accurate approach to replace potentially tedious and error-prone UI actions.
Taking the task \textit{``setting of an alarm''} as an example, an agent could directly tell the app the user's intent to set an alarm, rather than acting on behalf of the user to locate and enter the alarm app, tap to set the alarm, adjust the time, and then confirm.
The latter approach complicates the agent and reduces efficiency. Therefore, our work abstracts intent invocation as function calling and implements operations on Android through function calling, rather than through UI interactions.

\subsection{LLMs for Function Calling}
\begin{figure*}[htb]
    \centering
    \includegraphics[width=0.80\linewidth]{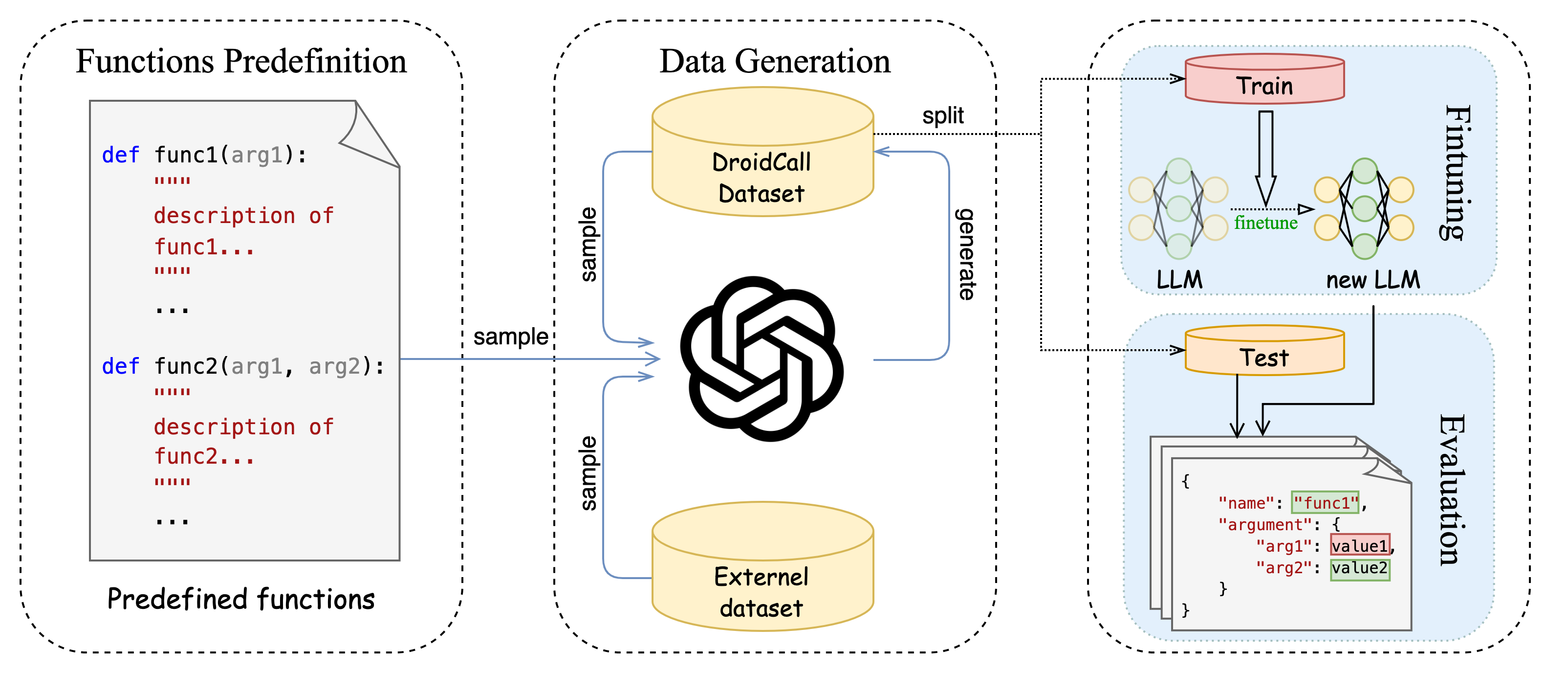}
    \caption{Workflow of \sys, which consist of three key phases:(1) Functions Predefinition; (2) Data Generation; (3) Finetuning and Evaluation.}
    \label{fig:workflows}
\end{figure*}

LLMs possess a certain level of reasoning capability, which enables them to make function calls when needed. Toolformer \cite{schick2024toolformer} is a pioneering work in this area; it attempts to teach LLMs to use tools during interactions with users. This work demonstrates the feasibility of LLMs performing function calls and provided a framework for subsequent research and development. To equip models with the capability of function calling, a substantial amount of data is often required. Self-Instruct \cite{selfinstruct} proves that it is possible to use LLMs like GPT to generate a significant volume of data for fine-tuning. Following the self-instruct paradigm, numerous efforts \cite{qin2023toolllm, tang2023toolalpaca, patil2023gorilla, kim2023llmcompiler} have been made to generate vast amount of function calling data for the purpose of fine-tuning models. This kind of approach makes it possible to equip models with function-calling capabilities by fine-tuning open-source models. Some work like APIGen \cite{liu2024apigenautomatedpipelinegenerating}, ShortcutsBench \cite{shen2024shortcutsbench} and ToolACE \cite{liu2024toolace} focus on dataset construction.
To effectively utilize diverse data sources, AgentOhana \cite{zhang2024agentohana} standardizes the data format and designs a training pipeline for effective agent learning.
In our work, we construct a reusable and highly customizable data generation pipeline. Unlike some other work that generate general function calling data, we focus on generating function-calling data related to Android intent invocation, which makes it possible to achieve a better performance on edge than GPT-4o.
We also provide a set of simple and easy-to-use methods for fine-tuning and evaluation. TinyAgent \cite{erdogan2024tinyagent} and Octopus \cite{chen2024octopus} are similar to our work; they both implement function-calling agents on mobile devices. However, TinyAgent focuses on operations on Mac, while Octopus requires adjustments to the model architecture (by expanding the vocabulary).
None of them provide code for data generation or model fine-tuning.

\section{\sys Dataset and Workflow}

In this section, we introduce the overall workflow of \sys, which comprises three key phases as shown in Figure~\ref{fig:workflows}: Function Predefinition, Data generation, Finetuning and Evaluation. In $\S$\ref{subsec:intent}, we first introduce Android intent, a key mechanism of Android. Based on the common intents in Android, we manually predefine 24 functions that can assist users in performing some common operations on Android. In $\S$\ref{subsec:data_gen}, we detail our method for generating the \sys dataset, the first open-sourced dataset for Android intent invocation.
Our method requires minimal human supervision and can be easily extended.
In $\S$\ref{subsec:finetune}, we describe how we fine-tune LLMs and evaluate their performance.
$\S$\ref{subsec:put_together} shows an end-to-end demonstration of device control using fine-tuned LLMs with \sys.

\subsection{Collecting Android Intents}\label{subsec:intent}

\begin{figure*}[htbp]
	\centering  
	\subfigure[Implicit intent \cite{implicit-intent}]{
        \label{fig:implicit-intent}
		\includegraphics[width=0.45\linewidth]{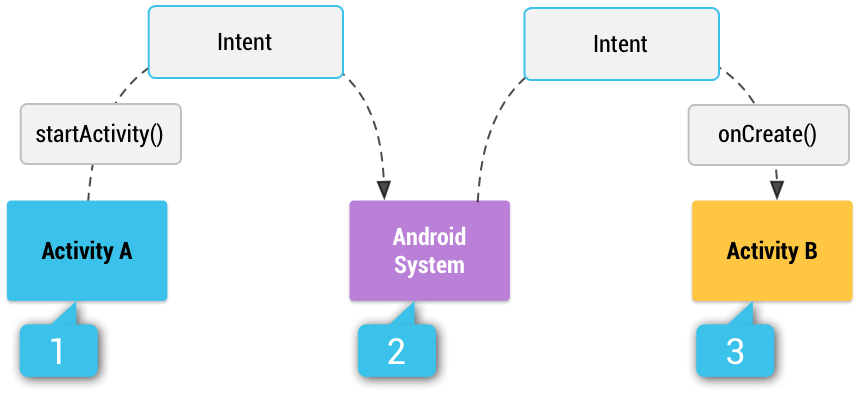}}
	\quad 
	\subfigure[Set alarm]{
		\label{fig:intent example}
		\includegraphics[width=0.45\linewidth]{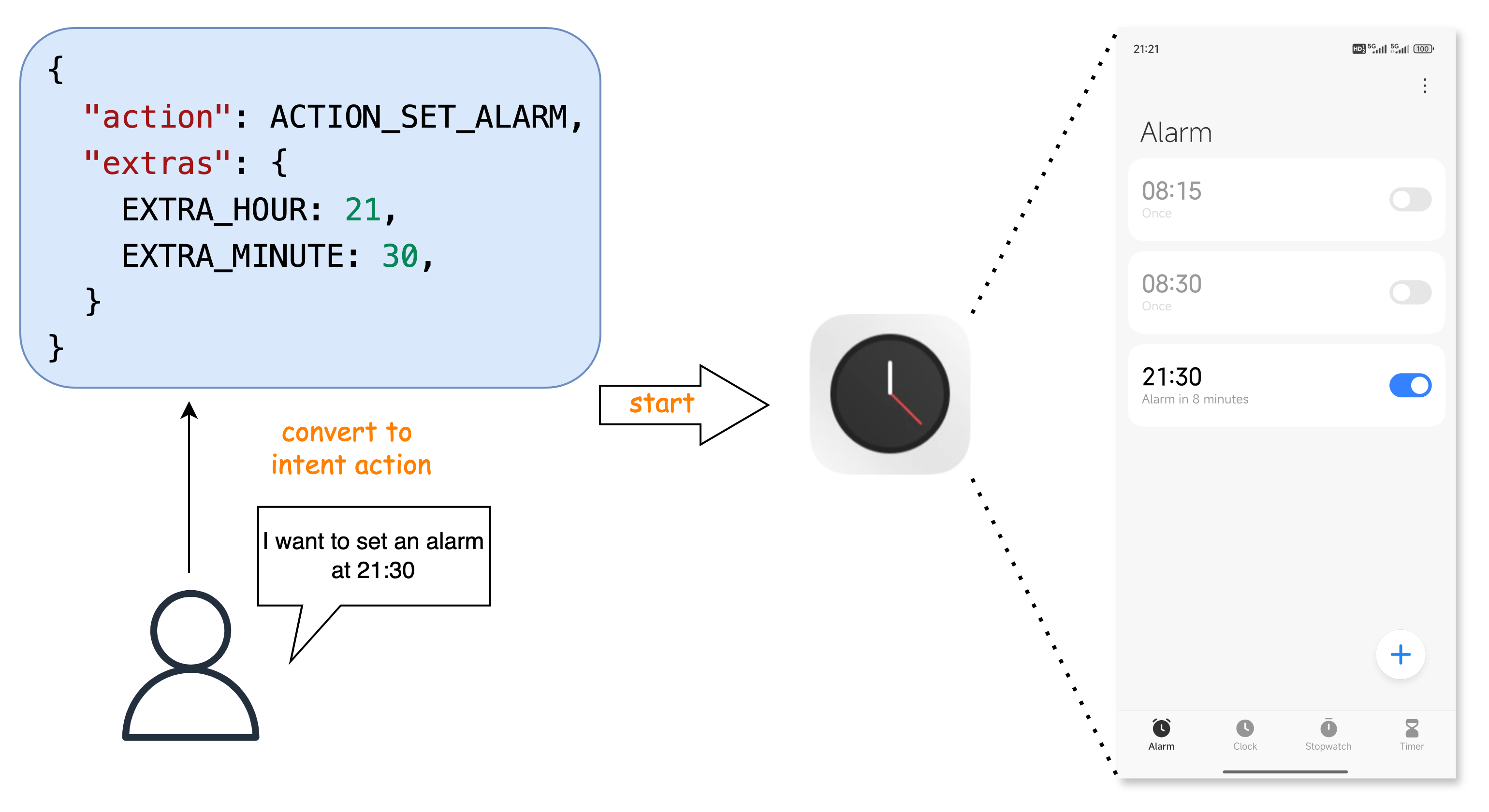}}
	\caption{(a) shows how implicit intent works in Android; (b) shows an example to help user set an alarm with implicit intent.
    }
	\label{fig:pretraining-training-loss}
\end{figure*}

In Android development, an intent is a fundamental messaging object used to request an action from another app component. Despite its simplicity, the intent system plays a crucial role in facilitating communication between components in an Android application. It acts as a glue that connects individual components such as activities, services, broadcast receivers, and content providers. Android intent can be divided into two categories:
\begin{itemize}
    \item \textbf{Explicit Intents} specify the exact component to start by providing the fully qualified class name. This type is typically used for internal communication within the app. One can start an activity in response to a user action or start a service to do some background work.
    
    \item \textbf{Implicit Intents.} Unlike explicit intents, implicit intents do not directly specify the target component. Instead, they declare a general action to perform as shown in Figure~\ref{fig:implicit-intent}, which allows any app component that can handle this action to respond. This makes implicit intents especially useful for interacting with components from different applications. For instance, if an app needs to capture a photo, it can issue an implicit intent to utilize the camera application installed on the device.
\end{itemize}

The primary motivation to build \sys is to enable models to perform function calling on Android devices for common operations, thereby enhancing user assistance capabilities. Android intent, as its name suggests, serves as a mechanism to express user ``intentions'' and trigger corresponding activities to fulfill these intentions. We identify implicit intents as the optimal choice for implementing common mobile operations, as they effectively articulate user intentions while maximizing the utilization of system resources to meet user requirements. Figure~\ref{fig:intent example} shows an example of setting an alarm using implicit intent.

To construct the \sys dataset, we review the Android official documentation~\cite{common-intents} and select frequently-used intents. These intents will be encapsulated into a set of functions that serve as the foundation for the \sys dataset generation. These functions encompass a broad range of common Android operations, including but not limited to alarm configuration, email composition, web searching.

\subsection{Dataset Generation}\label{subsec:data_gen}

\begin{figure}[htbp]
    \centering
    \includegraphics[width=1\linewidth]{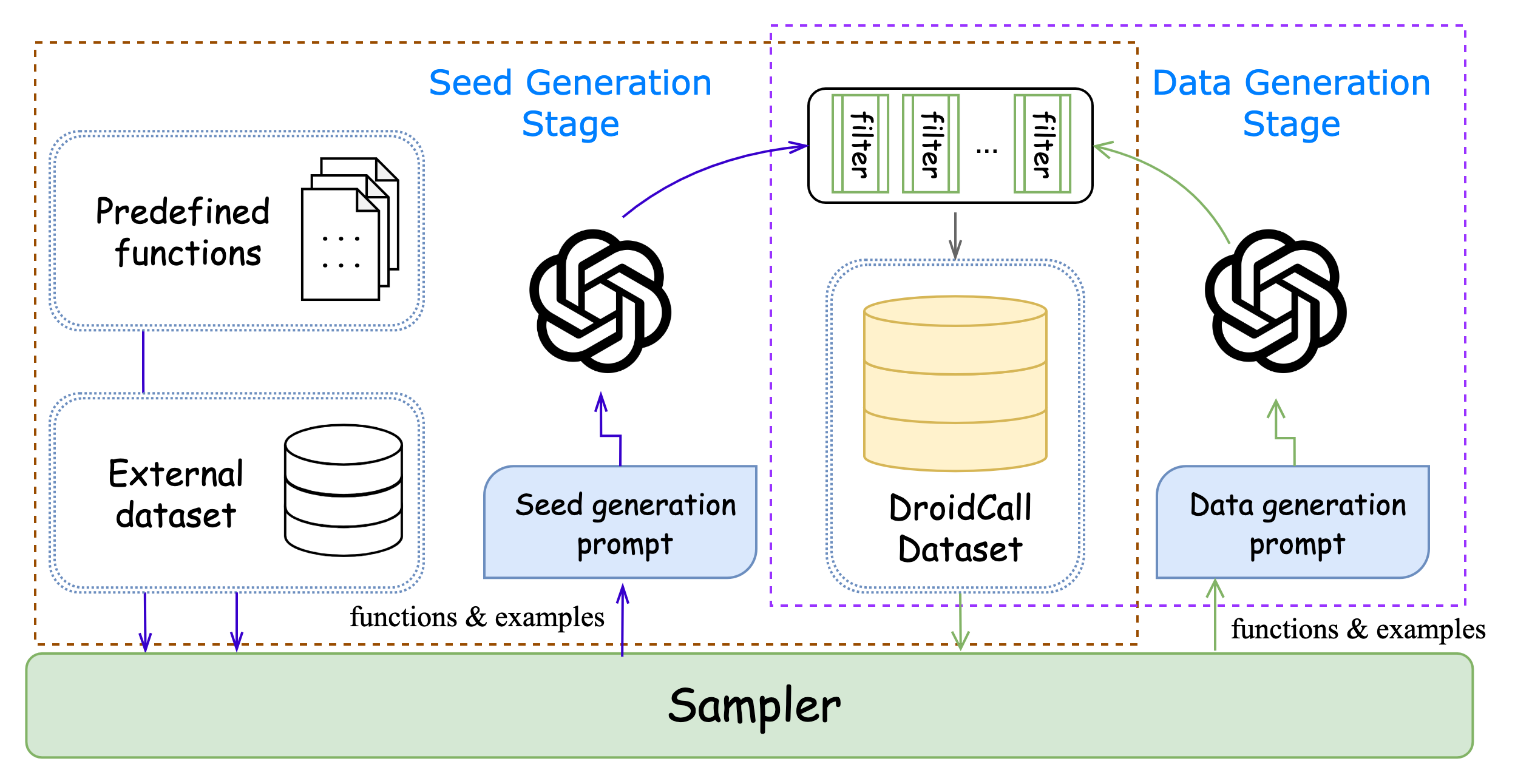}
    \caption{Details of data generation in \sys. To avoid manually creating seed data, \sys initially samples examples from an external dataset to generate its first set of data. Subsequently, the data is used as seed data to continuously generate new data, thereby eliminating the need for laborious manual work. All the generated data will go through a set of customized filters to ensure the correctness of data formats and the diversity of the data.}
    \label{fig:data_generation}
\end{figure}

In this section, we present a detailed description of the \sys dataset generation process. We first introduce the key components utilized in data generation: the \textit{sampler}, \textit{collector}, \textit{LLM} and \textit{filter} components. Subsequently, we elaborate on the critical phases of data generation: function predefinition, seed data generation, and data generation. The entire dataset generation process leverages GPT-4-turbo as the underlying language model. Figure~\ref{fig:data_generation} shows an overall workflow of data generation.

\subsubsection{Key Components of Generation Pipeline}

\textit{Sampler}, \textit{LLM}, \textit{Filter}, and \textit{Collector} are essential components of the data generation pipeline.

\textbf{Sampler.} A sampler is the component capable of taking multiple data sources as input (such as lists, jsonl files, etc.), sampling the required data from each data source according to a specific sampling strategy, and organizing the sampled data into a particular format as intended by the user for output. 

\textbf{LLM}. LLM serves as the engine for data generation. We employ the self-instruct \cite{selfinstruct} paradigm for data generation, which involves integrating data sampled by the sampler into specific prompt templates and then handing it over to the LLM for data generation. Therefore, the LLM is the core component of data generation, and its capabilities directly impact the quality of the generated data. In this paper, GPT-4-turbo is utilized as the LLM for data generation.

\textbf{Filter.} A filter is used to process the output from the LLM after it has been generated, such as extracting structured data from the output of LLMs, discarding data that does not meet the required format, and eliminating data that is highly similar to existing data. In the framework, the LLM can be processed by a series of custom filters, offering a high degree of flexibility.

\textbf{Collector.} A collector is the component designed to coordinate and utilize the aforementioned three components, acting as the manager of the entire data generation pipeline. After obtaining data from the sampler, the collector integrates the data into specified prompt templates, generates raw data through the LLM, processes the data generated by the LLM with a series of filters, and ultimately collects the results obtained.

\subsubsection{Functions Predefinition}

The automated extraction of intents from the Android Open Source Project (AOSP) \cite{AOSP} source code is a complex endeavor due to the multitude of intents and the dynamic nature of the Android platform. Given the complexities involved in the automated extraction of intents, our methodology diverges from such approaches.

We have predefined 24 functions that cover common operations on Android and utilize common intents within these functions for their specific implementations. Subsequently, we will teach the LLM to operate Android by learning to use these predefined functions. These predefined functions act as an interface between the LLM and the intents, concealing the specific details of the intents from the LLM. This approach also circumvents the issue of different versions of Android causing the LLM's learned intent knowledge to become obsolete. Once the LLM has learned these functions, we only need to implement them on different versions of Android, and the LLM will be able to do the intents invocation on different versions of Android through the functions without needing to know the specifics of the intents. In particular, these functions can perform common operations on Android, which include:
\begin{itemize}
    \item \textbf{Scheduling Assistant}: Help users to set an alarm/timer, insert an event on calendar.
    \item \textbf{Contact Management}: Add contacts, make phone calls.
    \item \textbf{Common Operations}: Internet search, search on maps, open camera for taking photos or recording videos, open various settings.
    \item \textbf{Messaging Services}: Compose text messages or emails.
\end{itemize}

In our framework, the method for predefining functions is the same as that for defining ordinary Python functions. We only need to write out the function signatures and provide the Google-style docstrings \cite{GoogleDocstrings} for the function.
Subsequently, we can automatically extract structured information describing the function from the function signature and the docstring. The extracted data have the format shown in Listing~\ref{lst:function}.
\begin{lstlisting}[language=json, caption={Extracted function. ``returns'' field and ``example'' field are optional.}, label=lst:function,captionpos=b]
{
  "name": "func1",
  "description": "This function is ...",
  "arguments": {
    "arg1": {
        "description": "This arg is...", 
        "type": "<type>",
        "required": "true or false",
        "default": "<default_value>"
    },
    "arg2": ...
  },
  "returns": { 
    "type": "...",
    "description": "..."
  },
  "example": [...]
}
\end{lstlisting}

\subsubsection{Data generation}

We follow the self-instruct paradigm \cite{selfinstruct, alpaca} to build our data generation pipeline. We have two stages to generate data.

\textbf{Seed Generation Stage.} When leveraging LLMs for synthetic data generation, incorporating high-quality examples in the prompt is crucial. This guidance helps maintain the data's quality and aligns it closely with human standards. These examples are called seed. It is often the case that these seeds are manually written and verified, which can be very labor-intensive and time-consuming. To avoid the time-consuming and labor-intensive task of manually writing seed data, we automatically generate a series of seed data for the function before formally generating data. The specific method is to use existing function-calling datasets as examples to guide the LLM in generating seed data. Specifically, we sample data from xlam-function-calling-60k \cite{liu2024apigenautomatedpipelinegenerating} and prompt the LLM to generate user queries and calling examples for our predefined functions based on the given instances. The data generated from this process will serve as seed data for subsequent data generation stages. 

\textbf{Data Generation Stage.} During this stage, we directly employ the self-instruct paradigm for data generation. For a predefined function, denoted as \textit{func}, we have previously generated a series of data for \textit{func} in the seed generation stage. In this stage, we will extract several instances from these seed data to serve as examples for the LLM, enabling it to produce more user queries and \textit{func} calling examples. The format of generated data is shown in Listing~\ref{lst:data}.
\begin{lstlisting}[language=json, caption={An example of generated data}, label=lst:data,captionpos=b]
{
  "query": "user query here",
  "answers": [
    {
        "id": id,
        "name": "func_name",
        "arguments": {
            "arg1": "value1",
            ...
        }
    },
    ...
  ]
}
\end{lstlisting}

Regardless of the stage, the output from the LLM will be processed through several custom filters. In our setup, the following three filters were utilized in sequence.

\textbf{JsonExtractor.} This filter is designed to extract JSON data from the output of LLM. It has been observed that GPT4-turbo does not always strictly adhere to a specific format for output. To properly handle the output of LLM, this filter employs a syntax parser to extract JSON data from the output of LLM.

\textbf{FormatFilter.} To ensure the extracted JSON strictly match the format in Listing~\ref{lst:data}, we use \textit{FormatFilter} to filter out those not in the proper format.

\textbf{SimilarityFilter.} This filter is designed to address the issue of high data similarity and poor quality due to the LLM's tendency to generate similar examples. It works by tracking the user queries that have been generated and calculating the LCS ROUGE score \cite{lin-2004-rouge} for each new user query against the existing data. When the F-measure value exceeds 75\%, the data is filtered out.

In our experiment, we generate two types of function-calling data.
\begin{itemize}
    \item \textbf{Simple.} The user's query is simple and straightforward, requiring the use of a single function for one call. In this case, the sampler just sample a single predefined function for LLM to generate data.
    \begin{lstlisting}[language=json, caption={An example of simple call in which a simple invocation of ACTION\_SET\_ALARM can fulfill the user's requirement.}, label=lst:data,captionpos=b]
{
  "query": "Wake me up at 8:30",
  "answers": [
    {
      "id": 0,
      "name": "ACTION_SET_ALARM",
      "arguments": {
        "EXTRA_HOUR": 8,
        "EXTRA_MINUTE": 30
      }
    }
  ]
}
\end{lstlisting}

    \item \textbf{Complex.} The user's query is intricate and cannot be resolved with a single function call. Instead, it requires the combination of two or more functions to address the complexity of the request. In the process of generating this type of data, the sampler samples two to three functions from the predefined set and prompts the LLM to generate examples of complex function calls. 
    \begin{lstlisting}[language=json, caption={An example of complex call in which we need to call two functions to fulfill the user's requirement.}, label=lst:data,captionpos=b]
{
  "query": "Set a timer for 30 minutes and dial 123456",
  "answers": [
    {
      "id": 0,
      "name": "ACTION_SET_TIMER",
      "arguments": {
        "duration": "30 minutes"
      }
    },
    {
      "id": 1,
      "name": "dial",
      "arguments": {
        "phone_number": "123456"
      }
    }
  ]
}
\end{lstlisting}
    
\end{itemize}

Using the method described above, we generated the \sys dataset, which is made up of two parts: train and test. The train split contains 10,000 data entries, while the test split contains 200 data entries.
We provide all prompt templates used to create \sys in Appendix~\ref{sec:appendix-a}.

\subsection{Fine-tuning SLMs with \sys}\label{subsec:finetune}

\textbf{Models.}
We fine-tuned a series of SLMs using the \sys dataset, including PhoneLM-1.5B \cite{yi2024phonelmanefficientcapablesmall}, Qwen2.5-1.5B, Qwen2.5-3B \cite{qwen2,qwen2.5}, Llama3.2-1B, Llama3.2-3B \cite{dubey2024llama}, MiniCPM3-4B \cite{hu2024minicpmunveilingpotentialsmall}, Phi3.5-3.8B \cite{abdin2024phi3technicalreporthighly} and Gemma2-2B \cite{gemmateam2024gemma2improvingopen}.

\textbf{Modeling function-calling tasks.}
We regard function calling as an instruction following task, where the model's input consists of \textit{user query}, \textit{available function descriptions}, and \textit{task instructions}. The output of the model is a \textit{specific representation for calling a function}.

If a unified input-output format is designed for fine-tuning models from different vendors, there would be issues: different models use different formats instruction tuning. If we use a unified format for fine-tuning, this format may have a gap compared to the format used during the model's instruction tuning. This could potentially affect the model's performance when it comes to function calling. Most current models have undergone fine-tuning specifically for chat, which typically involve three roles: system, user, and assistant. So we can reuse model's own chat template to do function calling. Specifically, we put \textit{user query} and \textit{available function descriptions} in system prompt and user prompt and put the function calling result in assistant output. By adopting this approach, we can equip the model with the capability for function calling while avoiding a significant gap between the data used for fine-tuning and the knowledge the model has already acquired. 

\textbf{Setups.}
We formatted the \sys dataset into the chat format described above, resulting in 10K training samples. We then fine-tuned the model using LoRA \cite{hu2022lora}, with a LoRA rank of 8 and a LoRA alpha of 16. Additionally, we employed a linear learning rate scheduler, setting the learning rate to 1.41e-5 and the warmup ratio to 0.1. We train for 24 epoch and pick the best checkpoint. Details of prompt format are provided in Appendix~\ref{sec:appendix-b}.

\subsection{Putting It All Together}\label{subsec:put_together}

\begin{figure}[htbp]
    \centering
    \includegraphics[width=0.9\linewidth]{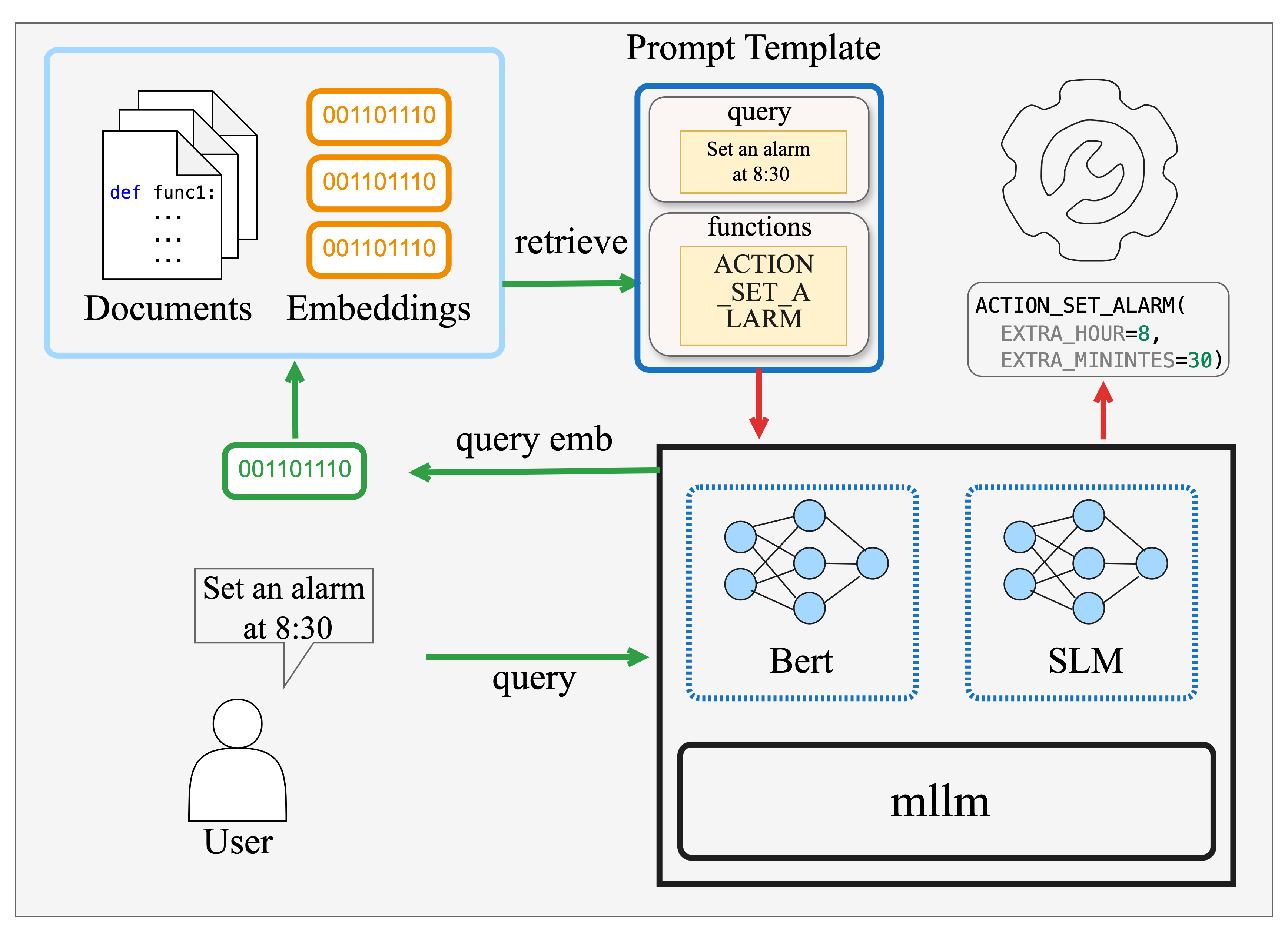}
    \caption{Design of our demo.}
    \label{fig:demo_design}
\end{figure}

Using the \sys dataset, we equip SLMs with certain capabilities for Android intent invocation. To verify its effectiveness, we developed an Android application. The design of our demo is shown in Figure~\ref{fig:demo_design}. This demo consists of two important components. One is the retriever, which is used to retrieve the most relevant functions. To implement this retriever, we utilized GTE \cite{GTE} to create word embeddings for the function descriptions and stored them in ObjectBox \cite{objectbox}, a vector database. When a user query arrives, we employ GTE for word embedding and retrieve the most relevant functions from ObjectBox, thus we have a simple and effective retriever. Another important component is a model capable of intent invocation, which takes in the user's query along with the functions retrieved by the retriever and outputs the function calls that can fulfill the user query. In our demo, we used PhoneLM-1.5B \cite{yi2024phonelmanefficientcapablesmall} fine-tuned on the \sys dataset as this model. It is worth noting that all of our model inference processes are completed on mobile phones. We utilized mllm \cite{yi2023mllm}, a fast and lightweight multimodal LLM inference engine designed for mobile and edge devices, to carry out the inference for both GTE and PhoneLM. Finally, we have a demo that can help us completing common tasks on Andrid devices.

Figure~\ref{fig:demo} illustrates an example of the end-to-end demo, in which the fine-tuned model understands the user's intent and assists users in adding an event to the calendar.

\section{Experiments}
\label{sec:experiments}

We first explored the impact of different prompt designs on model fine-tuning. After considering both the length of the prompts and the model's performance after fine-tuning, we selected an appropriate format for our prompts. Subsequently, through experimentation, we demonstrated that using the \sys dataset is more effective than using a general function calling dataset in scenarios aimed at Android intent invocation. Finally, we presented a series of results showcasing the effectiveness of models fine-tuned with the \sys dataset.

\textbf{Metrics.}
To quantitatively assess the efficacy of function calling within our model, we introduce two distinct metrics: \textit{Accuracy} and \textit{Soft Accuracy}.

\begin{itemize}
    \item \textit{Accuracy.} This metric evaluates the model's ability to precisely replicate the ground-truth function calls associated with a user query. A sample is deemed correct if and only if the model's output perfectly matches the ground truth in terms of both function identity and parameter values. Mathematically, Accuracy (\( Acc \)) is formalized as the ratio of the number of perfectly predicted samples (\( N_{\text{perfect}} \)) to the total number of samples (\( N_{\text{total}} \)):
    \[
    Acc = \frac{N_{\text{perfect}}}{N_{\text{total}}}
    \]

    \item \textit{Soft Accuracy.} This metric offers a nuanced evaluation of the model's performance, especially when it produces function calls that are partially correct. For each function call, a score is assigned based on the proportion of accurately predicted parameters (\( P_{\text{correct}} \)) relative to the total number of parameters (\( P_{\text{total}} \)). Soft Accuracy (\( A_{\text{soft}} \)) is then computed as the mean of these scores across all function calls:
    \[
    Acc_{\text{soft}} = \frac{1}{F} \sum_{i=1}^{F} \frac{P_{\text{correct},i}}{P_{\text{total},i}}
    \]
    where \( F \) denotes the total number of function calls.
\end{itemize}

It is important to note that the parameters of some functions are not straightforward to compare directly for correctness, such as parameters like \textit{title} or \textit{subject}. Semantic consistency is sufficient for these parameters; they do not need to match exactly to be considered correct. For such parameters, we employ models from the RoBERTa \cite{liu2019robertarobustlyoptimizedbert} series to compare semantic similarity. If the similarity exceeds a set threshold, it is considered correct. We set the threshold to 0.75.

We use the 200 data entries from the test split of \sys to evaluate SLMs.
It is worth noting that in real-world applications, we need to use a retriever to retrieve the functions that are likely to be used. But in our work, we are not focused on the retriever. So when testing the $Acc$ and $Acc_{soft}$, we use a fake retriever that always retrieves the ground-truth functions.

\subsection{Effect of Different Prompts}
\label{subsec:prompt}

\begin{figure*}[htpb] 
	\centering  
	
	\subfigure[Accuracy of \textit{Qwen2.5-1.5B-Instruct} on different prompts]{
        \label{fig:prompt_format}
		\includegraphics[width=0.45\linewidth]{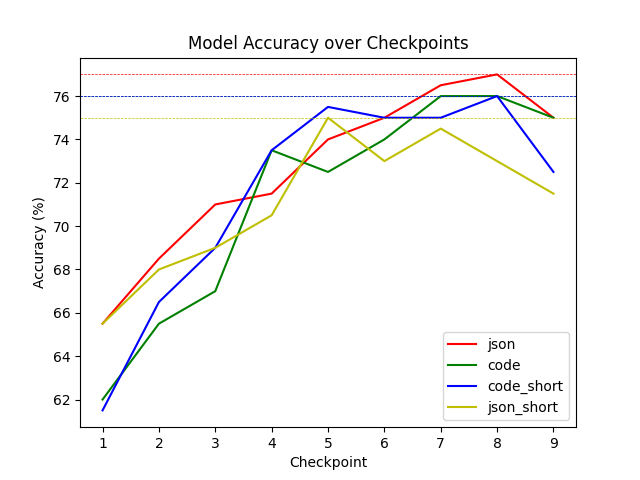}}
	\quad 
	\subfigure[Different models fine-tuned on different datasets]{
		\label{fig:dataset-comparison}
		\includegraphics[width=0.45\linewidth]{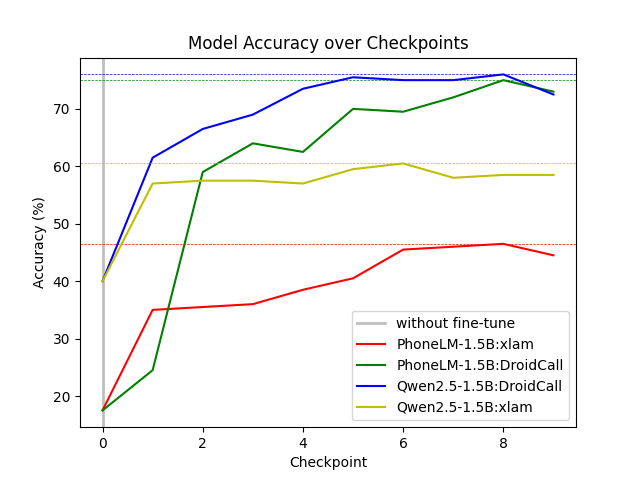}}
	\caption{Figure~\ref{fig:prompt_format} illustrates the performance of \textit{Qwen2.5-1.5B-Instruct} after fine-tuning under different prompt formats. Figure~\ref{fig:dataset-comparison} shows the performance of \textit{PhoneLM-1.5B} and \textit{Qwen2.5-1.5B-Instruct} after finetuning on different datasets.
}

	\label{fig:pretraining-training-loss}
\end{figure*}

\begin{table}[htbp]
    \centering
    \begin{tabular}{c|c}
        \hline
        \textbf{Prompt} & \textbf{Average Number of Tokens} \\ \hline
        code\_short & 645.195 \\ \hline
        json\_short & 950.340 \\ \hline
        code & 931.555 \\ \hline
        json & 1367.905 \\ \hline
    \end{tabular}
    \caption{Average number of tokens of different prompts.}
    \label{tab:tokens}
\end{table}

\begin{table*}[t]
\centering
\begin{tabular}{@{}p{0.2\textwidth}p{0.1\textwidth}@{}cccccc@{}}
\toprule
\multirow{2}{*}{\textbf{Model}} & \multirow{2}{*}{\textbf{Size}} & \multicolumn{2}{c}{\textbf{Zero-Shot}} & \multicolumn{2}{c}{\textbf{Few-Shot}} & \multicolumn{2}{c}{\textbf{Fine-Tuning}} \\ \cmidrule(lr){3-4} \cmidrule(lr){5-6} \cmidrule(l){7-8}
 & & $Acc$ & $Acc_{soft}$ & $Acc$ & $Acc_{soft}$ & $Acc$ & $Acc_{soft}$ \\ \midrule
PhoneLM-1.5B & 1.5B & 17.5 & 17.5 & 55.5 & 62.8 & 75 & 86.1 \\
Qwen2.5-1.5B-Instruct & 1.5B & 61 & 76.6 & 64.5 & 81 & 76 & 90.3 \\
Qwen2.5-3B-Instruct & 3B & 62 & 79.4 & 71 & 86.1 & 83 & 93.5 \\
Qwen2.5-Coder-1.5B & 1.5B & 42.5 & 48.8 & 65.5 & 81.6 & 82 & 93.2 \\
Gemma2-2B-it & 2B & 59 & 77.2 & 67.5 & 83.7 & 85 & 93.9 \\ 
Phi-3.5-mini-instruct & 3.8B & 62 & 77.8 & 67.5 & 82.1 & 83.5 & 93.8 \\ 
MiniCPM3-4B & 4B & 67 & 84.3 & 75 & 89.6 & 74.5 & 82.3 \\ 
Llama3.2-1B-Instruct & 1B & 31.5 & 37.7 & 60.5 & 76.3 & 75.5 & 87.3 \\ 
Llama3.2-3B-Instruct & 3B & 66.5 & 79.8 & 72 & 87.2 & 82 & 92.7 \\ 
GPT-4o &  & 77 & 89.1 & 80.5 & 91.5 &  & \\ 
GPT-4o-mini &  & 71.5 & 86.6 & 76 & 88.6 &  &  \\ 
\bottomrule
\end{tabular}
\caption{Evaluation of different models}
\label{tab:eval}
\end{table*}

In $\S$~\ref{subsec:finetune}, we mentioned that the model input includes several key components: \textit{user query}, \textit{available function descriptions}, and \textit{task instructions}. The model output is a \textit{specific representation for calling a function}. The \textit{user query} is provided by the user and is beyond our control. However, we can design the remaining three parts and observe how models perform after fine-tuning using different designs.

\textbf{json} A minimalist and straightforward design is to directly use JSON data as \textit{available function descriptions} and a \textit{specific representation for calling a function}. We opt for JSON formatting due to its simplicity.

\textbf{code} Another approach is to leverage the prevalence of code data in large language models' pre-training. We hypothesize that using \textit{docstrings} as \textit{available function descriptions} and adopting a \textit{Python function call} format for the \textit{specific representation for calling a function} may yield superior results. This is because such data closely resembles the coding examples encountered during pre-training, potentially enhancing the model's comprehension and performance. 

\textbf{short} In the above two formats, we provided a detailed description of the tasks the model is required to complete as the \textit{task instructions}. However, this approach significantly increases the length of the prompt. We posit that for models undergoing fine-tuning, \textit{task instructions} may not be essential. Through the fine-tuning process, models can learn to perform the function calling task without the need for explicit \textit{task instructions} in the prompt. Consequently, we experimented with removing the \textit{task instructions} from the previous formats, which we denote as \textit{json\_short} and \textit{code\_short}.

We experimented with fine-tuning the Qwen2.5-1.5B-Instruct model using four different types of prompts mentioned above. We selected nine checkpoints throughout the entire fine-tuning process to test for accuracy, with the results shown in Figure~\ref{fig:prompt_format}. As can be observed from the figure, the final outcome indicates that the \textit{json} format performed slightly better. However, the upward trend among all four formats is consistent, and the \textit{code\_short} format is essentially on par with the \textit{code} and \textit{json} formats. Additionally, we tested the average of input tokens for the model under four different prompt formats on the \sys test dataset, as shown in Table~\ref{tab:tokens}. The \textit{code\_short} format has a significantly smaller number of tokens compared to the other prompts. After comprehensive consideration, we ultimately chose the \textit{code\_short} format for subsequent fine-tuning experiments.

\subsection{Effectiveness of \sys}

To verify that the \sys dataset can achieve better results in the task of controlling Android phones through Android Intent invocation, we compared the performance of the Qwen2.5-1.5B-Instruct and PhoneLM-1.5B models after fine-tuning on the \sys dataset and xlam-function-calling-60k \cite{liu2024apigenautomatedpipelinegenerating}, a general function calling dataset.

To eliminate the influence of prompt design, we formatted both the xlam-function-calling-60k and \sys datasets using the \textit{code\_short} format. The xlam-function-calling-60k dataset comprises 60k data points, while \sys contains 10k. To ensure an equivalent number of training data instances, we trained the model for 4 epochs on the xlam-function-calling-60k dataset and for 24 epochs on \sys. We selected 9 checkpoints for testing, and the results are presented in Figure~\ref{fig:dataset-comparison}. Note that the 0th checkpoint in the figure represents the model's performance when directly evaluated with the \textit{code} format of prompts before any fine-tuning took place.

From the experimental results, we can observe that, regardless of the dataset used, accuracy improves with the fine-tuning process. However, the model fine-tuned with the xlam-function-calling-60k dataset quickly reaches a plateau. In contrast, the improvement brought by using the \sys dataset is significantly more substantial.

It is evident that when a model is required to perform a specific task, a dataset constructed for that task, such as \sys, can yield better results compared to a general-purpose dataset. Furthermore, from Figure~\ref{fig:dataset-comparison}, we can discern that initially, Qwen's capabilities are significantly higher than PhoneLM's. However, by the end of the fine-tuning process, PhoneLM's performance is on par with Qwen's. We speculate that initially, PhoneLM's Supervised Fine-Tuning (SFT) and alignment were not as effective as Qwen's, preventing it from leveraging its pre-trained knowledge efficiently. The \sys dataset, however, aids the model in learning to utilize its pre-trained knowledge to control Android devices. Since the pre-trained knowledge of both PhoneLM and Qwen is comparable, they eventually reach a similar level of performance.

\subsection{Performance of Different SLMs}

To test the Android intent invocation capabilities of some existing SLMs tailored for the edge scenario and further verify the effectiveness of \sys, we tested the $Acc$ and $Acc_{soft}$ of a few models under three conditions: zero-shot, few-shot, and after fine-tuning. When testing the models' zero-shot and few-shot performance, we used \textit{json} format prompts for both, as the \textit{json-short} and \textit{code-short} formats lack \textit{task instructions}, which prevents the model from finishing the task. In comparison, the \textit{json} format has been found to be more effective than the \textit{code} format.

The experimental outcomes, as depicted in Table~\ref{tab:eval}, provide a comprehensive overview of the Android intent invocation capabilities across various models. Judging from the zero-shot results, there is a significant performance variation among different models. The zero-shot scenario is a critical test of a model's ability to complete tasks based on instructions without having seen relevant examples. We believe the primary reason for the differences in zero-shot performance among models lies in the effectiveness of their SFT and alignment. These training stages determine whether the model can develop strong instruction-following capabilities. It is also observable that all models exhibit improved performance under few shots. We credit the performance boost to the models' improved use of their knowledge from pretraining. 

After fine-tuning with \sys, there is a significant improvement in the model's performance. Additionally, during inference, the model only requires a prompt that essentially consists of the \textit{user query} and \textit{available function descriptions}, which greatly reduces the prompt length compared to the zero-shot and few-shot scenarios.

\section{Conclusion}

In this paper, we introduce \sys, a novel dataset specifically engineered to enhance the Android intent invocation capabilities of LLMs. Our approach diverged from conventional cloud-based models, focusing instead on on-device deployment to address privacy concerns inherent in mobile environments. In our work, we (1) build a highly customizable and reusable data generation pipeline, (2) construct \sys, a first-of-its-kind open-sourced dataset for Android intent invocation based on the pipeline, (3) fine-tune a series of models tailored for edge devices, enabling them to approach or even surpass the performance of GPT-4o in the specific task of intent invocation, (4) implement an end-to-end demo with mllm. Our work demonstrates the potential applications of small models on the edge. We have open-sourced all the code of the data generation, fine-tuning, and evaluation.

\nocite{langley00}

\bibliography{ref}
\bibliographystyle{icml2021}


\appendix
\onecolumn
\section{Data Generation Prompts}
\label{sec:appendix-a}

At the beginning of data generation, we first generate seed data. The prompt used to generate seed is shown as following:

\begin{mdframed}
I need your help to generate some function calling datasets. I will provide you with a tool description, and you need to generate queries and corresponding answers based on this tool, i.e., the answers that call the tool to resolve the user's query. Here are my requirements:

1. For queries, try to use different vocabulary and syntax to ensure query diversity. Queries can be long or short, complex or concise. In short, try not to generate similar queries; I want to ensure query diversity. \\
2. The language of the queries should be as diverse as possible. This means a query can be a command, a question, or a request with detailed descriptions, etc. \\
3. The generated queries should cover all possible uses of the tool as much as possible, meaning the coverage of various parameters should be comprehensive, ensuring the tool can be used to complete various forms of work. \\
4. The generated queries should be solvable using the given tools. \\
5. For the queries you generate, you should provide answers using the tool, i.e., give the tool used and the values for each parameter. \\
6. When providing parameters, if a parameter has required=False, you may omit its value. \\
7. The generated data must be presented in the format given in my example. \\
8. The parameter values generated with function call generated must be values that can be inferred from the user's query; YOU CANNOT FABRICATE PARAMETERS THAT CANNOT BE OBTAINED FROM THE USER'S REQUEST. \\
9. Attach each answer with an id starting from 0. And if a tool should use the respone from another tool, you can reference it using \#id, where id is the id of the tool. \\

following are some examples: \\
\$examples

Now I will give you a tool, and you help me generate 15 query-answer pairs. \\
REMEMBER TO GENERATE THE RESULT IN JSON FORMAT LIKE THE EXAMPLE ABOVE
REMEMBER NOT TO FABRICATE PARAMETERS FOR TOOLS. PARAMETERS SHOULD BE INFERED FROM USER QUERY. \\
tool: \$tool
\end{mdframed}

In the prompt above, \$examples will be replace by random samples sampled from xlam-function-calling-60k \cite{liu2024apigenautomatedpipelinegenerating}. Below is an example:

\begin{mdframed}
\begin{verbatim}
tool: {
    "name: "...",
    "description": "...",
    "arguments": {
        ...
    }
}
response: {
    "query": "...",
    "answers": [
        {
            ...
        }
    ]
}
\end{verbatim}
\end{mdframed}
\$tools will be replace by json formatted predefined function, below is an example:
\begin{mdframed}
\begin{verbatim}
tool: {
    "name": "ACTION_SET_ALARM",
    "description": "...".
    "arguments": {
        ...
    }
}
\end{verbatim}
\end{mdframed}

After seed generation stage, we will use another prompt to continuously generate data. Prompt is shown as following:
\begin{mdframed}
I need your help to generate some function calling datasets. I will provide you with a tool description and some example data for you. 
You need to generate queries and corresponding answers based on this tool, i.e., the answers that call the tool to resolve the user's query. Here are my requirements:

1. For queries, try to use different vocabulary and syntax to ensure query diversity. Queries can be long or short, complex or concise. In short, try not to generate similar queries; I want to ensure query diversity.\\
2. The language of the queries should be as diverse as possible. This means a query can be a command, a question, or a request with detailed descriptions, etc.\\
3. The generated queries should cover all possible uses of the tool as much as possible, meaning the coverage of various parameters should be comprehensive, ensuring the tool can be used to complete various forms of work.\\
4. The generated queries should be solvable using the given tools.\\
5. For the queries you generate, you should provide answers using the tool, i.e., give the tool used and the values for each parameter.\\
6. When providing parameters, if a parameter has required=False, it is not necessary to provide its value.\\
7. The query-answer pairs should cover as many possible uses of the tool as possible.\\
8. The generated data must be presented in the format given in my example.\\
9. The parameter values generated with function call generated must be values that can be inferred from the user's query; YOU CANNOT FABRICATE PARAMETERS THAT CANNOT BE OBTAINED FROM THE USER'S REQUEST.\\

following are tool I provided and some examples of query-answer pairs:
tool: \$tool
examples: \$examples

Now please help me generate 40 query-answer pairs.
REMEMBER TO GENERATE THE RESULT IN JSON FORMAT LIKE THE EXAMPLE ABOVE
REMEMBER NOT TO FABRICATE PARAMETERS FOR TOOLS. PARAMETERS SHOULD BE INFERED FROM USER QUERY.
\end{mdframed}
\$tool will be replaced by the json format of predefined functions shown early. \$examples is the data sampled from the seed data generated previously.

When generating data of complex call, we slightly modify the prompt shown above. The seed generation prompt is shown below:
\begin{mdframed}
I need your help to generate some function calling datasets. I will provide you with a tool description, and you need to generate queries and corresponding answers based on this tool, i.e., the answers that call the tool to resolve the user's query. Here are my requirements:

1. For queries, try to use different vocabulary and syntax to ensure query diversity. Queries can be long or short, complex or concise. In short, try not to generate similar queries; I want to ensure query diversity.\\
2. The language of the queries should be as diverse as possible. This means a query can be a command, a question, or a request with detailed descriptions, etc.\\
3. The generated queries should cover all possible uses of the tool as much as possible, meaning the coverage of various parameters should be comprehensive, ensuring the tool can be used to complete various forms of work.\\
4. The generated queries should be solvable using the given tools.\\
5. For the queries you generate, you should provide answers using the tool, i.e., give the tool used and the values for each parameter.\\
6. When providing parameters, if a parameter has required=False, you may omit its value.\\
7. The generated data must be presented in the format given in my example.\\
8. THE PARAMETER VALUES GENERATED WITH FUNCTION CALL GENERATED MUST BE VALUES THAT CAN BE INFERRED FROM THE USER'S QUERY; YOU CANNOT FABRICATE PARAMETERS THAT CANNOT BE OBTAINED FROM THE USER'S REQUEST.\\
9. THE GENERATED QUERY SHOULD CONTAIN ENOUGH INFOMATION SO THAT YOU COULD CORRECTLY GENERATE PARAMETER USED BY THE TOOLS. THIS IS
 ALSO TO GUARANTEE THAT YOU DON'T FABRICATE PARAMETERS.\\
10. You should use all the tools I provided to generate the query and answer. It means that you should generate a query that needs to use all the tools I provided to solve, and remember to provider an answer that uses all the tools to solve the query.\\
11. You can use the same tool multiple times in a single query to ensure the query diversity.\\
12. Attach each answer with an id starting from 0. And if a tool should use the respone from another tool, you can reference it using \#id, where id is the id of the tool.\\
13. Generate data of nested function calls if possible. i.e., the argument of a function call is the response of another function call.\\

following are some examples:\\
\$examples

Now I will give you a tool, and you help me generate 15 query-answer pairs.
REMEMBER TO GENERATE THE RESULT IN JSON FORMAT LIKE THE EXAMPLE ABOVE AND PUT IT IN A JSON LIST.\\
REMEMBER YOU SHOULD USE ALL THE TOOLS AT ONE QUERY AND SOLVE IT WITH ALL TOOLS, AND GENERATE NESTED CALL IF POSSIBLE.\\
REMEMBER NOT TO FABRICATE PARAMETERS FOR TOOLS. PARAMETERS SHOULD BE INFERED FROM USER QUERY.\\
tools: \\
\$tools
\end{mdframed}
Prompt for continuously generating complex function calling data is:
\begin{mdframed}
I need your help to generate some function calling datasets. I will provide you with a tool description, and you need to generate queries and corresponding answers based on this tool, i.e., the answers that call the tool to resolve the user's query. Here are my requirements:

1. For queries, try to use different vocabulary and syntax to ensure query diversity. Queries can be long or short, complex or concise. In short, try not to generate similar queries; I want to ensure query diversity.\\
2. The language of the queries should be as diverse as possible. This means a query can be a command, a question, or a request with detailed descriptions, etc.\\
3. The generated queries should cover all possible uses of the tool as much as possible, meaning the coverage of various parameters should be comprehensive, ensuring the tool can be used to complete various forms of work.\\
4. The generated queries should be solvable using the given tools.\\
5. For the queries you generate, you should provide answers using the tool, i.e., give the tool used and the values for each parameter.\\
6. When providing parameters, if a parameter has required=False, you may omit its value.\\
7. The generated data must be presented in the format given in my example.\\
8. THE PARAMETER VALUES GENERATED WITH FUNCTION CALL GENERATED MUST BE VALUES THAT CAN BE INFERRED FROM THE USER'S QUERY; YOU CANNOT FABRICATE PARAMETERS THAT CANNOT BE OBTAINED FROM THE USER'S REQUEST.\\
9. THE GENERATED QUERY SHOULD CONTAIN ENOUGH INFOMATION SO THAT YOU COULD CORRECTLY GENERATE PARAMETER USED BY THE TOOLS. THIS IS
 ALSO TO GUARANTEE THAT YOU DON'T FABRICATE PARAMETERS.\\
10. You should use all the tools I provided to generate the query and answer. It means that you should generate a query that needs to use all the tools I provided to solve, and remember to provider an answer that uses all the tools to solve the query.\\
11. You can use the same tool multiple times in a single query to ensure the query diversity.\\
12. Attach each answer with an id starting from 0. And if a tool should use the respone from another tool, you can reference it using \#id, where id is the id of the tool.\\
13. Generate data of nested function calls if possible. i.e., the argument of a function call is the response of another function call.\\

Now I will give you some tools and some example data of query-answer pairs using these tools. 
Please help me generate 40 query-answer pairs.
tools: \$tools\\
examples: \$examples

REMEMBER TO GENERATE THE RESULT IN JSON FORMAT LIKE THE EXAMPLE ABOVE AND PUT IT IN A JSON LIST.\\
REMEMBER YOU SHOULD USE ALL THE TOOLS AT ONE QUERY AND SOLVE IT WITH ALL TOOLS, AND GENERATE NESTED CALL IF POSSIBLE.\\
REMEMBER NOT TO FABRICATE PARAMETERS FOR TOOLS. PARAMETERS SHOULD BE INFERED FROM USER QUERY.
\end{mdframed}

\section{Function Calling Prompts}
\label{sec:appendix-b}

In $\S$~\ref{subsec:prompt}, we've mentioned that we have tested 4 format of prompt: \textit{json}, \textit{code}, \textit{json\_short} and \textit{code\_short}. To unify our fine-tuning, we use chat to do function calling thus we only need to design the part of system, user and assistant using chat template.

In \textit{json} or \textit{code} format, the system prompt would be:
\begin{mdframed}
You are an expert in composing functions. You are given a query and a set of possible functions. 
Based on the query, you will need to make one or more function calls to achieve the purpose. 
If none of the function can be used, point it out. If the given question lacks the parameters required by the function,
also point it out. Remember you should not use functions that is not suitable for the query and only return the function call in tools call sections. 
\end{mdframed}

in \textit{json\_short} or \textit{code\_short} the system prompt would be:
\begin{mdframed}
You are an expert in composing functions. 
\end{mdframed}
The user part of \textit{json} or \textit{code} is:
\begin{mdframed}
Here is a list of functions that you can invoke:\\
\$functions

Should you decide to return the function call(s), Put it in the format of\\ 
\$format\_description

\$example

If there is a way to achieve the purpose using the given functions, please provide the function call(s) in the above format.
REMEMBER TO ONLY RETURN THE FUNCTION CALLS LIKE THE EXAMPLE ABOVE, NO OTHER INFORMATION SHOULD BE RETURNED.

Now my query is: \$user\_query
\end{mdframed}
\$functions is the functions descriptions provided by retriever, in \textit{code} or \textit{code\_short} format, it would be like:
\begin{mdframed}
\begin{verbatim}
Name:
    send_email
Description:
    Compose and send an email with optional attachments.

This function allows the user to compose an email with various options,
including multiple recipients, CC, BCC, and file attachments.
Args:
    to (List[str]): ...
    subject (str): ...
    ...
Returns:
    None
Example:
    # Send an email with a content URI attachment
send_email(
    to=["recipient@example.com"],
    subject="Document",
    body="Please find the attached document.",
    attachments=...
)
\end{verbatim}
\end{mdframed}
In \textit{json} or \textit{json\_short}, functions would be describe directly in json format as shown in Listing~\ref{lst:function}.

\$format\_description in the prompt will be replace by detailed output format the model should follow. In \textit{json} it will be:
\begin{mdframed}
\begin{verbatim}
[
    {
      "id": 0,
      "name": "func0",
      "arguments": {
          "arg1": "value1",
          "arg2": "value2",
          ...
      }
    },
    {
      "id": 1,
      "name": "func1",
      "arguments": {
          "arg1": "value1",
          "arg2": "value2",
          ...
      }
    },
    ...
]
If an argument is a response from a previous function call, 
you can reference it in the following way like the argument 
value of arg2 in func1:
[
    {
      "id": 0,
      "name": "func0",
      "arguments": {
          "arg1": "value1",
          "arg2": "value2",
          ...
      }
    },
    {
      "id": 1,
      "name": "func1",
      "arguments": {
          "arg1": "value1",
          "arg2": "#0",
          ...
      }
    },
    ...
]
This means that the value of arg2 in func1 is the return 
value from func0 (#0 means the response from the function call with id 0).
\end{verbatim}
\end{mdframed}
In \textit{code} format this will be
\begin{mdframed}
\begin{verbatim}
result1 = func0(arg1="value1", arg2="value2", ...)
result2 = func1(arg1="value1", arg2=result1, ...)
...
You can do nested function calling in the following way:
result1 = func0(arg1="value1", arg2="value2", ...)
result2 = func1(arg1="value1", arg2=result1, ...)
...
This means that the value of arg2 in func1 is the return value from func0.
\end{verbatim}
\end{mdframed}
\$example in the prompt is used to test few-shot performance of a model.

The user prompt of \textit{json\_short} or \textit{code\_short} is much simpler withou \textit{task instructions}:
\begin{mdframed}
Here is a list of functions:
\$functions

Now my query is: \$user\_query
\end{mdframed}

In \textit{code} or \textit{code\_short} format the assistant output would be:
\begin{mdframed}
\begin{verbatim}
<sep>result1 = func0(arg1="value1", arg2="value2", ...)
result2 = func1(arg1="value1", arg2=result1, ...)</sep>
\end{verbatim}
\end{mdframed}
where $<sep>$ and $</sep>$ can be any seperator set before fine-tuning.

In \textit{json} or \textit{json\_short} format the assistant output would be:
\begin{mdframed}
\begin{verbatim}
[
    {
      "id": 0,
      "name": "func0",
      "arguments": {
          "arg1": "value1",
          "arg2": "value2",
          ...
      }
    },
    ...
]
\end{verbatim}
\end{mdframed}

\twocolumn


\end{document}